\documentclass[12pt]{article}
\usepackage[style = apa,
            uniquename=false,
            maxbibnames=20,
            backend=biber]{biblatex}
\usepackage{newtxtext,newtxmath}
\usepackage{graphicx}
\usepackage{parskip}
\usepackage{amsmath}
\usepackage{tabularx}
\usepackage{graphicx}
\usepackage{adjustbox}
\usepackage{url}
\usepackage{siunitx}
\usepackage{setspace}
\usepackage{geometry}
\usepackage{authblk}
\AtEveryBibitem{\clearfield{note}}

\title{Camera trap classification with deep learning under ground truth uncertainty}
\author[1$\dagger$]{Leonard Hockerts}
\author[1*$\dagger$]{Peter S. Stewart}
\author[2]{Sarthak Arora}
\author[1]{Tiffany J. Vlaar}
\affil[1]{School of Mathematics \& Statistics, University of Glasgow, Glasgow, G12 8QQ, United Kingdom.}
\affil[2]{Collaborative Earth, Berkeley, California, United States}
\affil[*]{Corresponding author: peter.s.stewart@glasgow.ac.uk}
\affil[$^\dagger$]{Joint first author}
\date{}

\begin{document}
\maketitle

\textbf{ORCIDs:} Leonard Hockerts (\url{https://orcid.org/0009-0003-0175-2546}); Peter S. Stewart (\url{https://orcid.org/0000-0001-6338-6279}); Sarthak Arora (\url{https://orcid.org/0009-0002-5769-4747}); Tiffany J. Vlaar (\url{https://orcid.org/0000-0002-0885-2393})

\textbf{Keywords}: Camera trap images, Class imbalance, Computer vision, Convolutional neural networks, Deep learning, Ground truth uncertainty, Pre-training

\textbf{Running head:} Camera traps and ground truth uncertainty

\textbf{Word count}: 7995 words

\textbf{Data availability statement}: We used open-access camera trap image datasets available at \url{https://doi.org/10.5061/dryad.5pt92} and \url{https://doi.org/10.5281/zenodo.14389489}. All analysis code is available at: \url{https://github.com/Leonard-H/prickly-pear-cnn}. 

\textbf{Author contributions}: PSS and TJV conceived and led the project, including the methodological design and analyses; LH and SA contributed to the methodological design and conducted the analyses; LH and PSS created the figures; PSS led the writing of the manuscript, with substantial contributions from TJV; all authors contributed critically to the drafts and gave final approval for publication. 

\textbf{Conflict of interest statement}: The authors have no conflict of interest to declare. 

\textbf{Acknowledgements}: For the purpose of open access, the authors have applied a Creative Commons Attribution (CC BY) licence to any Author Accepted Manuscript version arising from this submission. LH was supported by the EPSRC summer vacation internship. 

\newpage

\begin{abstract}

    \begin{enumerate}
        \item Supervised deep learning methods enable the rapid processing of large quantities of ecological image data, but depend on a costly and time-consuming annotation process. Consequently, training labels are commonly derived from volunteer citizen science projects. However, disagreement among volunteers introduces uncertainty in the ``ground truth" data that are assumed to be correct for model training and validation. The consequences of this ground truth uncertainty are poorly understood, yet potentially important across the broad range of applications where data have multiple labels.
        
        \item Using two datasets containing camera trap images with associated volunteer and expert classifications, we investigated the effects of training under higher ground truth uncertainty on model performance across a range of image difficulties, species-level classification accuracy, and generalisation across datasets. We also explored the interaction between ground truth uncertainty and pre-training, as well as class imbalance.
        
         \item Training under higher ground truth uncertainty improved overall test accuracy, particularly for images that were more difficult for human volunteers. Species-level accuracy also generally improved, including for several classes with relatively poor volunteer accuracy. However, generalisation to a different dataset did not improve. The benefits of ground truth uncertainty were enhanced by pre-training on ImageNet. Additionally, pre-training substantially reduced the number of training epochs required; further reductions in computational cost, but not gains in accuracy, resulted from additional pre-training on other camera trap images. With unbalanced training data, we still observed a clear benefit of increased ground truth uncertainty for overall accuracy, especially on difficult images. However, class imbalance affected species-level classification performance, improving accuracy for common species, reducing rare species accuracy, and changing patterns of misclassification to more closely resemble mistakes made by human volunteers. 

         \item Our findings have implications for the application of deep learning across ecological image types with multiple labels. Practitioners can improve classification accuracy, especially on difficult examples, by including moderate levels of label disagreement during training and using models pre-trained on general image data. In addition to improving the use of citizen science-derived labels in model training, our study suggests avenues for more effectively integrating human and deep learning classifications in combined workflows. 

    \end{enumerate}
\end{abstract}

\newpage

\section{Introduction}

Deep learning is transforming ecological research by enabling the rapid processing of vast quantities of image data, which would otherwise take researchers months or years to manually classify \parencite{christin_applications_2019, borowiec_deep_2022, pichler_machine_2023}. These approaches have been used to classify animals in camera trap images (\textit{e.g.,} \cite{Beery19, norouzzadeh2021}), plants in leaf photographs collected in the field (\textit{e.g.,} \cite{rzanny_acquiring_2017}) and from herbarium records (\textit{e.g.,} \cite{shirai_development_2022}), and plankton and bacteria in microscope images \parencite{bachimanchi_deep-learning-powered_2024, al-jumaili_novel_2026}. Deep learning is also commonly applied to satellite imagery and other remote sensing data, for tasks such as counting animals \parencite{Bowler2020}, classifying land cover types (\textit{e.g.,} \cite{mashraqi_hybrid_2025}) and monitoring large-scale changes over time (\textit{e.g.,} \cite{sefrin_deep_2021}). Additionally, deep learning has been applied to spectrogram images to classify species from acoustic data \parencite{lauha_domain-specific_2022}. These applications support fundamental ecological research in an increasingly data-rich field, and can facilitate rapid processing of data in applied conservation settings where timely action may be vital \parencite{christin_applications_2019, borowiec_deep_2022, pichler_machine_2023}. 

In image classification tasks, convolutional neural networks (CNNs; \cite{convnets, resnet}) are the dominant paradigm. However, the use of supervised learning approaches depends on a costly annotation process to generate training examples \parencite{norouzzadeh2021}. To mitigate this cost, it is common to use training image labels derived from volunteer citizen science projects \parencite{willi2019, green20, dickinson12, pocock18}. To improve the accuracy of this labelling process it is typical for each image to be classified by multiple volunteers, with a label being accepted when the proportion of volunteers in agreement passes a given threshold. To retain images for which there is no unanimous agreement -- there are usually significant differences amongst volunteer annotators' skillset, experience, interests, and behaviour that affect classification accuracy \parencite{johnston, Kosmala2016, Welinder2010} -- a threshold below 100\% can be used. The consequence of lowering this threshold is uncertainty in the “ground truth” data (i.e., image labels which are assumed to be correct, and which are used to train and validate the model; \cite{Bowler2020}).

Ground truth uncertainty arising from disagreements among volunteers has several potential consequences for the training of deep learning models. First, because volunteers are expected to disagree more frequently on images which are more difficult (\textit{e.g.,} where images are poor quality, visibility is poor or partially obscured, the class is intrinsically difficult to identify or may be confused with similar classes), training on these data allows the model to gain experience with more difficult examples. This is potentially important for addressing poor model performance on challenging images which are commonly obtained in ecology, but which remain under-represented in many datasets which are used to develop deep learning models \parencite{kitzes_integrating_2026}. In addition, retaining uncertain examples maximises the quantity of training data available. However, this also increases the risk of introducing incorrect training labels, which may damage model performance by training the model to associate the example with the wrong class. For instance, \cite{bevan_deep_2026} showed that incorrect labels can impact the accuracy of ecological metrics (species richness, occupancy, and activity) derived from camera trap data. However, as highlighted by the authors, since their study design applied incorrect labels uniformly across their dataset, their approach fails to capture realistic aspects of ground truth uncertainty such as the association of incorrect labels with certain classes or more difficult images. Consequently, the overall effect of ground truth uncertainty on the accuracy of classifications produced by deep learning models, and how this effect may vary among examples of different difficulty and across different classes, is not well understood. 

Ground truth uncertainty also has the potential to interact with two other key issues in the application of deep learning to ecological image data: generalisation to new contexts, and class imbalance. Classification models often fail to generalise well to new contexts or environments (\cite{Beery2018, shepley_automated_2021}), due to overfitting to the background or other irrelevant features. For example, \cite{sharpe25} found that Conservation AI's UK Mammal model \parencite{fergus_harnessing_2024} was substantially less accurate than expected when applied to their camera trap data. This poses a challenge for ecologists who wish to deploy pre-trained models on their own data, particularly if they do not have the computational resources or expertise to train their own model. As images with higher ground truth uncertainty may represent less typical examples of a class or background features (\textit{e.g.,} due to low visibility), it is possible that training on these images may improve performance on images from new contexts.

Class imbalance refers to an imbalance in the relative frequency of the examples for each class (\textit{e.g.,} species) in the training data. This is ubiquitous in ecological image datasets, for reasons including variation in species' abundance and detectability. Class imbalance represents a challenge for deep learning models because performance on abundant classes can dominate the loss function, incentivising the model to improve only on these classes and potentially harming performance on the rarer classes (\cite{prince_understanding_2024} p.73). Various approaches have been used to mitigate class imbalance \parencite{Shwartz2023}, including data resampling and data augmentation schemes (\textit{e.g.,} Trivial Augment; \cite{trivialaugment}). Ground truth uncertainty is expected to interact with class imbalance because some classes are intrinsically more likely to be uncertain – for example, a familiar and visually distinctive species like the elephant (\textit{Loxodonta africana}) may be less likely to be misidentified than a Grant’s gazelle (\textit{Nanger grantii}), which could be mistaken for another antelope species like impala (\textit{Aepyceros melampus}). Consequently, changing the volunteer agreement threshold can alter the degree of class imbalance in the dataset. 

Here, we leverage two datasets which comprise camera trap images with associated volunteer and expert labels to explore the effects of ground truth uncertainty on deep learning models for classifying ecological image data. Camera traps are one of the key sources of image data in ecology; they can be deployed in remote locations, are effective at monitoring rare and elusive animals, operate over all hours, and are often more cost-effective than other approaches \parencite{Bruce2025}. We address the following questions: 1) Does training under higher ground truth uncertainty improve performance on difficult images? 2) How is species-level classification accuracy affected by ground truth uncertainty?  3) Does training on more uncertain images improve generalisation between datasets? 4) Does ground truth uncertainty interact with class imbalance? Our findings offer insights into any setting where labels are provided by multiple annotators.

\section{Materials and Methods}
\subsection{Datasets}
We used two open-access datasets comprising camera trap images with associated volunteer and expert classifications: Prickly Pear Project Kenya (\cite{Stewart2025}; \url{https://doi.org/10.5281/zenodo.14389489}) and Snapshot Serengeti (\cite{swanson2015}; \url{https://doi.org/10.5061/dryad.5pt92}). 

Prickly Pear Project Kenya \parencite{Stewart2025} contains 186,861 images collected in Laikipia County, Kenya. The images contain $>$40 mammal species, ranging in size from dikdik (\textit{Madoqua} spp.) to the African savannah elephant (\textit{L. africana}). The images also contain $>$60 species of bird, mostly labelled as ``bird (other)''. Due to variation in species' abundance and detectability, the dataset exhibits a substantial level of class imbalance. The images were classified by 8290 volunteers who engaged with the project throught the Zooniverse platform (\url{https://www.zooniverse.org/}); the volunteers ranged widely in experience, from no prior experience to extensive experience classifying images on other Zooniverse projects. A relatively large number of images (n = 26,952) were classified by an expert.

Snapshot Serengeti \parencite{swanson2015} contains images of $>$40 species of mammal and bird taken in the Serengeti National Park, Tanzania.  
This dataset has been used in several other studies on the application of deep learning to camera trap images (\textit{e.g.,} \cite{norouzzadeh2021, norouzzadeh_automatically_2018, shepley_automated_2021, gomez_villa_towards_2017}) and as training data for widely used models such as MegaDetector \parencite{Beery19}. We used the subset for which consensus species classifications were available - these data consisted of 848,665 images from 319,915 capture events, of which 4149 events were classified by an expert. 

\subsection{Image and label preparation}
We focused on the most commonly observed species (17 species for Prickly Pear Project Kenya, 16 for Snapshot Serengeti) to ensure sufficient images for training and testing; we excluded images where the consensus classification (either the expert's classification, or the species that passed the volunteer agreement threshold) was not in the set of focal species. We also excluded images where the expert classifier recorded multiple species, or -- if no expert classification was present -- more than two volunteers recorded multiple species. Each Snapshot Serengeti capture event contained 1-12 images (median = 3, mean = 2.65); we used the middle image in each sequence and discarded the other images. For Prickly Pear Project Kenya, where images were retired after 12 classifications (\cite{Stewart2025}), we excluded images with fewer than 12 classifications. Some images had more than 12 classifications due to recycling completed images at the end of the project. For these cases, we used the first 12 classifications provided by different individuals. For Snapshot Serengeti, the maximum number of volunteer classifications was not standard across images; we therefore included all images with $\geq 10$ classifications. Following this process, Prickly Pear Project Kenya contained 128,996 images (118,736 volunteer-classified, 10,260 expert-classified), while Snapshot Serengeti contained 143,993 images (141,008 volunteer-classified, 2985 expert-classified). 

To remove image metadata and manufacturer logos we cropped all images (removing 66px from the bottom, 55px from the left, and 35px from the top for Prickly Pear Project Kenya; removing 100px from the bottom for Snapshot Serengeti).We then resized each image to 300px width and 160px height for Prickly Pear Project Kenya, and 300px width and 210px height for Snapshot Serengeti.

We derived training labels using a majority vote threshold-based approach, varying the threshold (100\%, 90\%, or 66\% agreement for label acceptance) to introduce different levels of ground truth uncertainty (see Section \ref{sec:experiments}). We used the expert-classified subset as the test set, using the expert's classification as the test label. 

\subsection{Model architecture and implementation}
We conducted our experiments using the ResNet (``residual network") architecture \parencite{resnet}. ResNet was one of the top-performing classification models on the ImageNet Large Scale Visual Recognition Challenge \parencite{imagenetchallenge} when it was originally developed \parencite{resnet}, and is often applied to ecological image data (\textit{e.g.,} \cite{tabak_machine_2019, norouzzadeh_automatically_2018, fergus_harnessing_2024}). We used ResNet-50, which has 50 hidden layers.

We used the Adam optimiser (\cite{adam}), using an initial learning rate of \num{5e-3} for models trained from scratch and \num{2e-3} for pre-trained models. We applied a learning rate scheduler with step size 5 and a multiplicative factor of 0.7, and used a batch size of 64. We trained all models until they reached a training accuracy of 0.97, recording the number of epochs required as a measure of the computational cost of training. For all models, we applied data augmentation consisting of random horizontal flips (p = 0.5), random rotations ($\pm 15^{\circ}$), and random affine transformations with up to 10\% translation in both directions. We replicated each experiment across three random seeds to help account for slight variations in model performance caused by stochasticity in the optimisation process, such as parameter initialisation (when training from scratch) and mini-batch sampling employed by the optimiser. All models were implemented using PyTorch \parencite{paszke_pytorch_2019} and run on NVIDIA RTX A6000 GPUs. 

\subsection{Exploring the effects of ground truth uncertainty}\label{sec:experiments}
We varied the training label agreement threshold  (100\%, 90\%, or 66\% agreement for label acceptance) to introduce different levels of uncertainty, and then assessed model performance on a test set comprised of expert-classified images. We primarily used the Prickly Pear Project Kenya dataset due to the larger pool of expert-classified images, but also replicated the experiments described in Sections \ref{sec:q1methods} and \ref{sec:q2methods} using Snapshot Serengeti. We addressed four main questions: 

\subsubsection{Does training under ground truth uncertainty improve performance on difficult images?}\label{sec:q1methods}
We trained models on balanced datasets created by randomly sampling 700 training images per species and using different label agreement thresholds (100\%, 90\%, or 66\%).  For the 90\% and 100\% volunteer agreement training datasets, some species were represented in fewer than 700 images. In these cases we duplicated a random subset of images until 700 was reached, using a fixed random seed to ensure that the duplicated images were consistent across experiments. No image was duplicated more than once. For the 90\% agreement Prickly Pear Project dataset, only the warthog (\textit{Phacochoerus africanus}; 641 images) required duplication. For the 100\% agreement dataset, the duplicated species were livestock (674 images), vervet monkey (\textit{Chlorocebus pygerythrus}; 585), camel (\textit{Camelus dromedarius}; 500), warthog (441), and helmeted guineafowl (\textit{Numida meleagris}; 351). For the Snapshot Serengeti 66\% dataset, only ostrich (\textit{Struthio camelus}418 images) required duplication. Ostrich also required duplication in the 90\% dataset (411 images), alongside baboon (\textit{Papio anubis}; 681). For the 100\% agreement dataset, the duplicated species were ostrich (380 images), baboon (496), Grant's gazelle (370), and impala (455). We trained models either from random initialisation, or pre-trained on ImageNet (\cite{imagenet}), a general image dataset containing over one million images representing 1000 classes.

\begin{figure}
    \centering
    \includegraphics[width=1\linewidth]{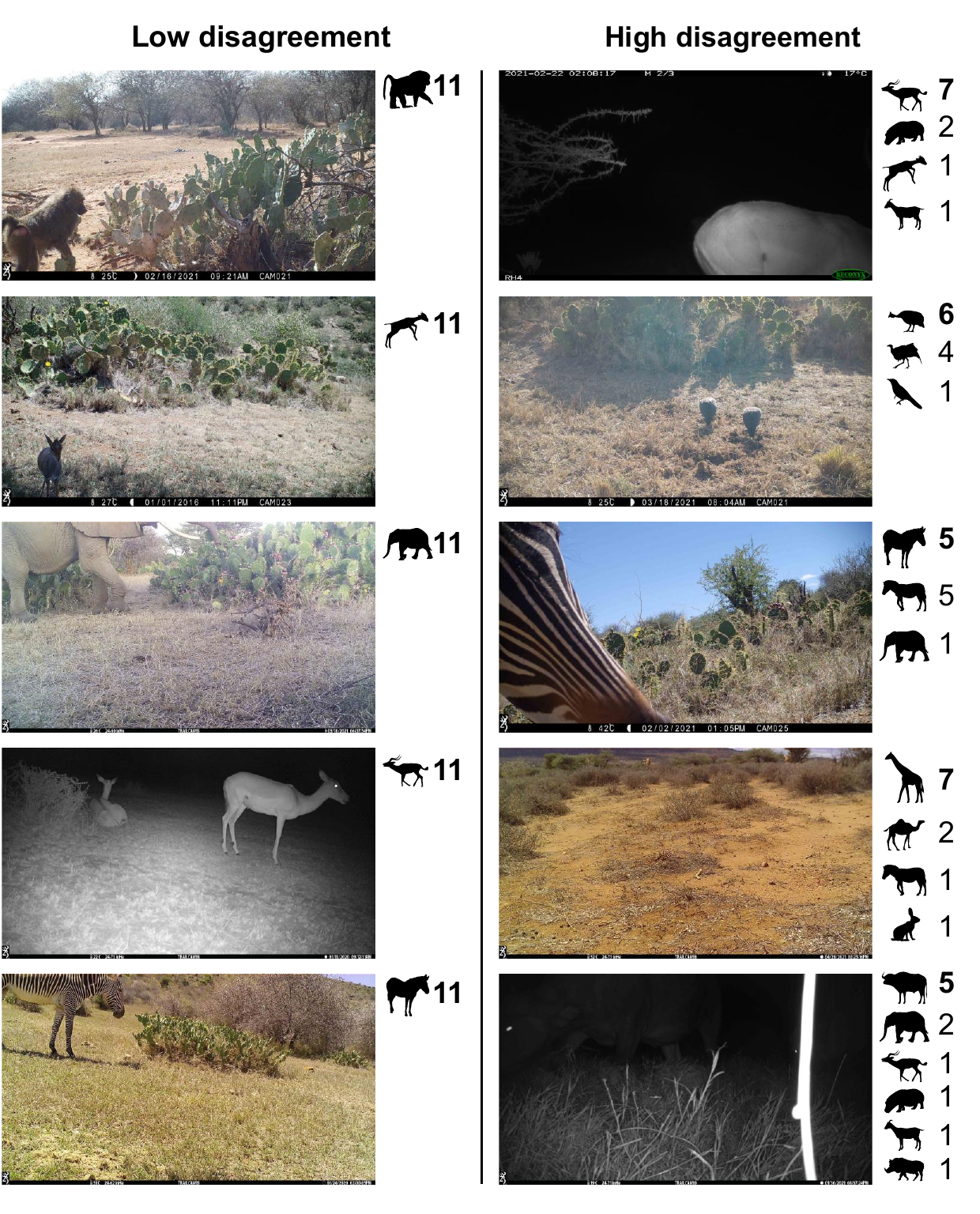}
    \caption{Example images with low (left) and high (right) volunteer disagreement from the Prickly Pear Project Kenya test set. Outlines with numbers represent the number of volunteer classifications for each species; the correct species is listed first in bold. For species names, see Fig. \ref{fig:species_heatmap_prickly_pear_balanced}.}
    \label{fig:image_examples}
\end{figure}

We then evaluated test accuracy across images representing a range of example difficulties. In more difficult examples, volunteers are expected to disagree more frequently on the true species classification. Therefore, we measured example difficulty by quantifying volunteer disagreement using the Shannon entropy \parencite{shannon} for each image in the test set: 
\begin{equation}
    H = -\sum\limits_{i=1}^N p_i \ \text{log}_2 \ p_i,\label{eq:entropy}
\end{equation}
where $p_i$ is the proportion of classifications for species $i$, and $N$ the total number of species categories. Higher Shannon Entropy values indicate more volunteer disagreement, and hence more difficult examples for humans to classify (Fig. \ref{fig:image_examples}). We examined how test accuracy varied across equal-width entropy bins. For Prickly Pear Project Kenya, five images with entropy $>1.5395$ were excluded due to the insufficient size of their bin. Similarly, for Snapshot Serengeti three images with entropy $>1.7364$ were excluded. We quantified overall test accuracy as the proportion of correct classifications:
\begin{equation}
    Accuracy = \frac{N_{correct}}{N_{total}}\label{eq:accuracy}
\end{equation}
where $N_{correct}$ and $N_{total}$ are the number of correct classifications and the total number of classifications respectively.

\subsubsection{How is species-level classification accuracy affected by ground truth uncertainty?}\label{sec:q2methods}
To quantify species-level performance, we used the true positive rate (TPR; \textit{i.e.,} recall or sensitivity):
\begin{equation}
    TPR = \frac{TP}{(TP+FN)} \label{eq:tpr}
\end{equation}
where TP and FN are the number of true positives and false negatives respectively. We quantified the accuracy of the human volunteers using the same metric for comparison.  

\subsubsection{Does training on more uncertain images improve generalisation between datasets?}\label{sec:q3methods}
To investigate whether training under higher ground truth uncertainty improved generalisation between datasets, we first trained models (either from scratch or following pre-training on ImageNet) on the balanced Snapshot Serengeti dataset, using either a 100\% or 66\% agreement threshold for label acceptance. We then fine-tuned the models to 0.97 training accuracy on the balanced Prickly Pear Project Kenya data, using either a 100\%, 90\%, or 66\% agreement threshold. Fine-tuning is commonly applied to enhance generalisation performance and reduce computational cost compared to training from scratch on the novel dataset \parencite{Yosinski2014,Radford2018,He2019}. Therefore, understanding whether conducting initial training under higher ground truth uncertainty affects model performance after the fine-tuning step is an important question for practitioners. We evaluated test accuracy across different entropy bins in the Prickly Pear Project Kenya test set as described above (Section \ref{sec:q1methods}). 

\subsubsection{Does ground truth uncertainty interact with class imbalance?}\label{sec:q4methods}
To investigate the interplay between ground truth uncertainty and class imbalance, we trained models on the Prickly Pear Project Kenya dataset without imposing a maximum number of images per class. As above, we trained models using 100\%, 90\%, and 66\% agreement thresholds; in contrast to the balanced setting, using different levels of agreement thresholds directly affects the number of training images and degree of imbalance. In this experiment, we pre-trained all models on ImageNet \parencite{imagenet}. We quantified the degree of imbalance in each training data by calculating the Shannon entropy (see Eq.\ref{eq:entropy}, where $p_i$ is the proportion of training images labelled as species $i$); higher entropy values represent more balanced data. We evaluated overall and species-level performance as described in Sections \ref{sec:q1methods} and \ref{sec:q2methods} using Eqs. \ref{eq:accuracy} and \ref{eq:tpr} respectively. 

\section{Results}
\subsubsection{Does training under ground truth uncertainty improve performance on difficult images?}
\begin{figure}
    \centering
    \includegraphics[width=1\linewidth]{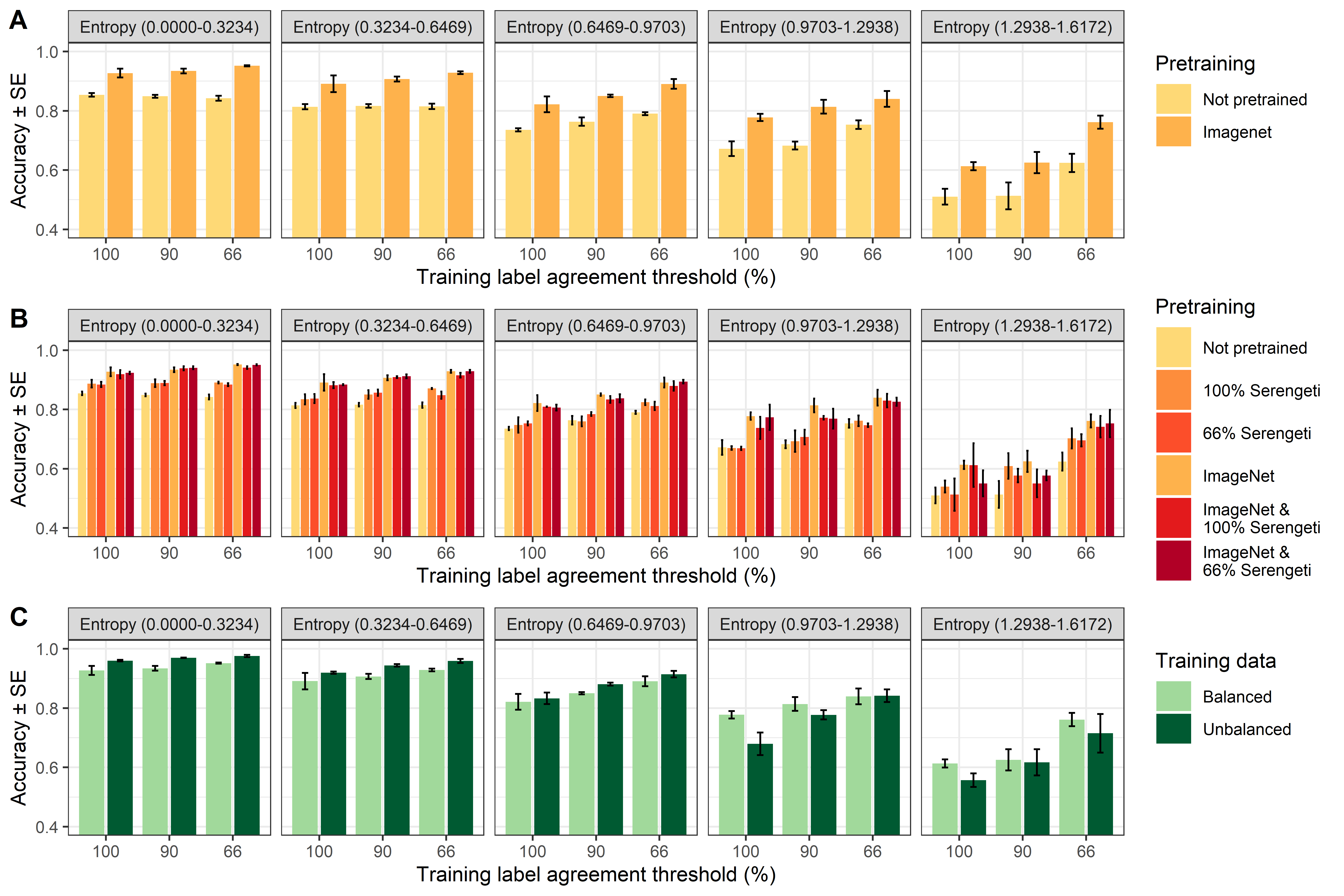}
    \caption{Test accuracy (proportion of correct classifications; mean $\pm$ standard error) for the Prickly Pear Project Kenya dataset under different volunteer agreement thresholds (\%) for label acceptance, for models which were \textbf{A)} either not pre-trained, or pre-trained on ImageNet, \textbf{B)} additionally pre-trained on Snapshot Serengeti images with either 100\% or 66\% volunteer agreement thresholds, and \textbf{C)} trained using a balanced or unbalanced dataset. Panels represent different levels of image difficulty as measured by the Shannon entropy of the volunteer classifications; higher entropy values (towards right) indicate more volunteer disagreement, and hence more difficult images. Sample sizes for entropy bins are (left to right) 6725, 2181, 1003, 255, and 92 images. Five images with entropy $>1.5395$ were excluded.
    }
    \label{fig:experiments_entropy}
\end{figure}

We found that as test image difficulty -- measured by the degree of volunteer disagreement (Shannon entropy) -- increased, the proportion of correct model classifications decreased (Fig. \ref{fig:experiments_entropy}A). However, the decrease in accuracy was smaller when the model was trained using a lower threshold for label agreement. Therefore, training under higher ground truth uncertainty improved model performance on images which were more difficult for human volunteers. 

Pre-training on ImageNet resulted in universally increased performance, relative to models that were not pre-trained (Fig. \ref{fig:experiments_entropy}A). Furthermore, in pre-trained models, the benefits of training under higher ground truth uncertainty were evident even at the lowest entropy bin, with the 66\% agreement threshold model slightly outperforming both the 90\% and 100\% threshold models (Fig. \ref{fig:experiments_entropy}A, left panel). By contrast, without pre-training the difference between label agreement thresholds only became apparent in the third entropy bin (Fig. \ref{fig:experiments_entropy}A, centre panel). Therefore, the benefits of training under higher ground truth uncertainty were amplified by pre-training on ImageNet. Overall, the best classification accuracy across all species was attained by pre-training on ImageNet and subsequently training under the highest level of ground truth uncertainty.  

When we repeated the experiment using Snapshot Serengeti, we again observed that model accuracy was lower for test images with higher volunteer disagreement (Fig. S1A). This decrease was smallest for the 66\% threshold, although the difference in performance between models was less pronounced than was observed for Prickly Pear Project Kenya. We also observed a clear benefit of pre-training on ImageNet across all label agreement thresholds and entropy levels. The best-performing model overall was pre-trained on ImageNet and subsequently trained using a 66\% threshold.

\begin{figure}
    \centering
    \includegraphics[width=0.75\linewidth]{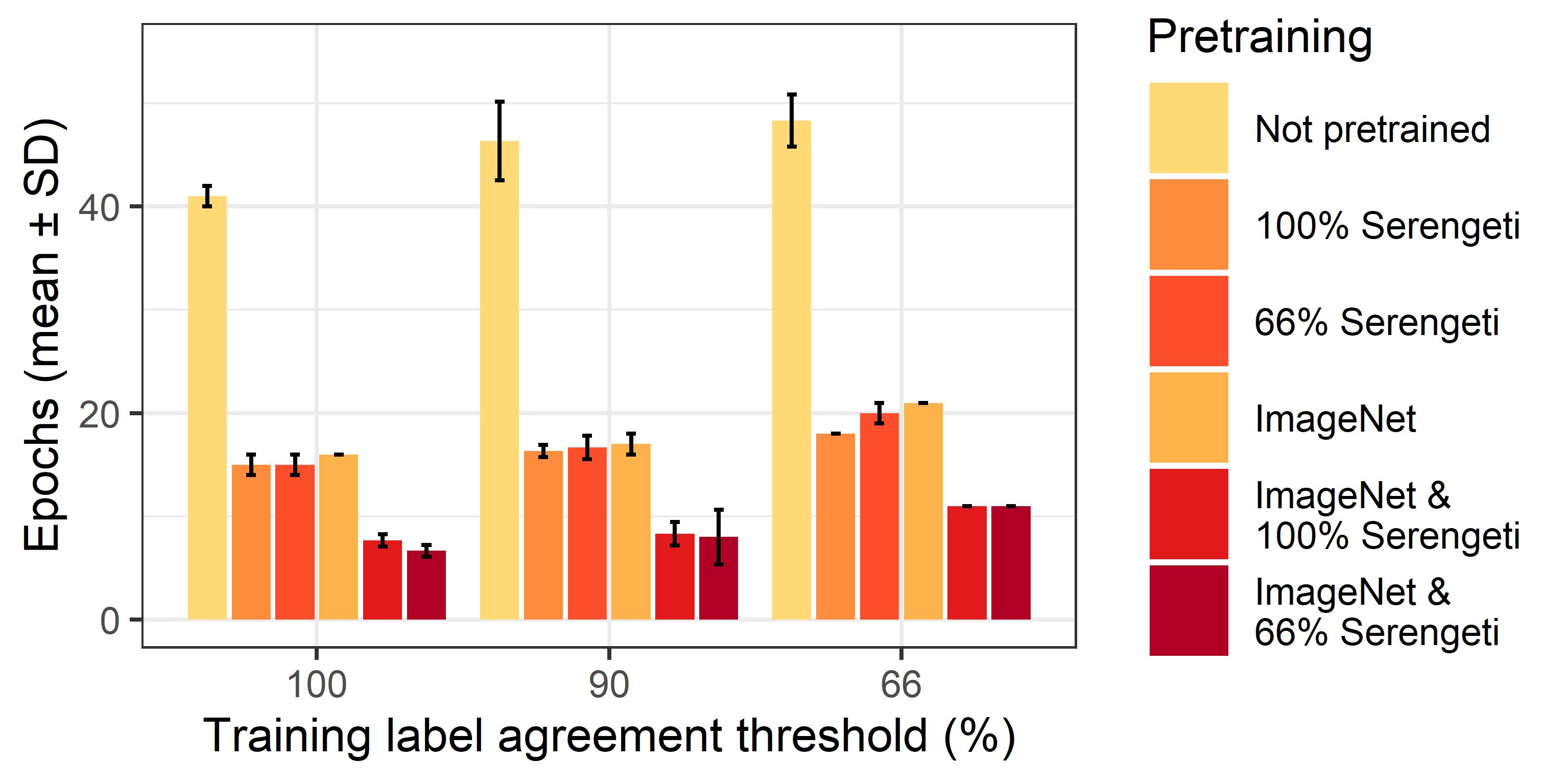}
    \caption{Number of epochs (mean $\pm$ standard deviation) taken to reach 0.97 training accuracy, for models trained with different label agreement thresholds (\%) and pre-training treatments. }
    \label{fig:experiments_epochs}
\end{figure}

\subsubsection{How is species-level classification accuracy affected by ground truth uncertainty?}
\begin{figure}
   \centering 
   \includegraphics[width=\linewidth] 
    {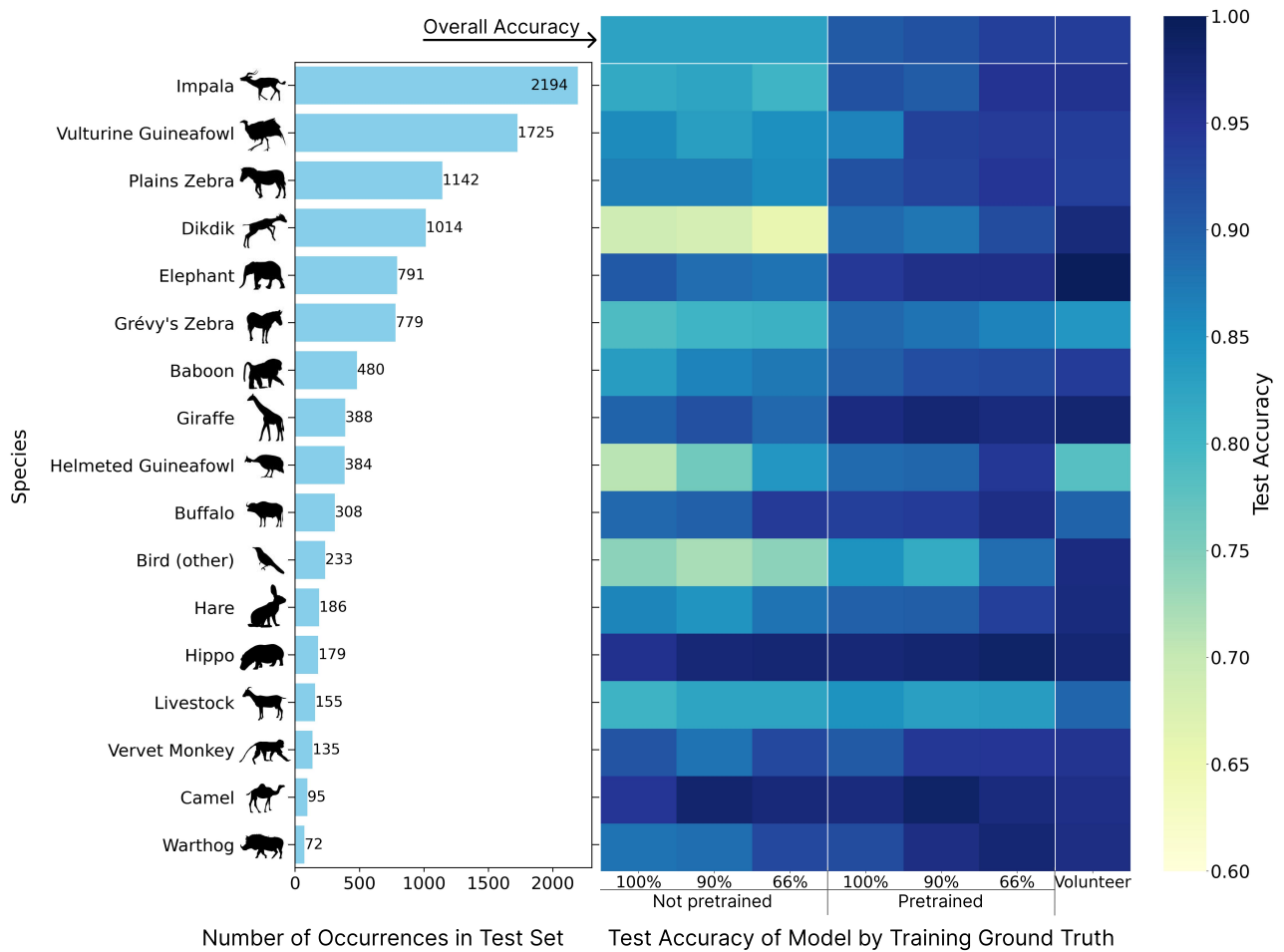}
        \caption{Species-level classification accuracy for the Prickly Pear Project Kenya Dataset, trained on a balanced dataset. \textbf{Left:} Number of images per species in test set.  \textbf{Right:} Test accuracy (true positive rate). Models were either trained from random initialization or pre-trained, and then trained on data with varying agreement thresholds ($\geq$66\%, 90\%, 100\%) for label acceptance. Volunteer accuracy is also displayed on the rightmost column. The top row displays overall test accuracy across all species.
    }   
    \label{fig:species_heatmap_prickly_pear_balanced}
\end{figure}

Training under higher ground truth uncertainty improved classification accuracy for several species; however, the strength of this effect varied among species, and depended on whether the model was pre-trained on ImageNet (Fig. \ref{fig:species_heatmap_prickly_pear_balanced}). The most pronounced effect was observed for helmeted guineafowl, the species with the lowest volunteer accuracy, across both pre-trained and non-pre-trained models. Clear accuracy gains were also observed for buffalo (\textit{Syncerus caffer}) -- which also displayed relatively low volunteer accuracy -- and warthog, regardless of pre-training. Models trained using 66\% agreement threshold data performed the best for vervet monkeys in both the pre-trained and non-pre-trained cases, although performance was also good for the 100\% threshold model without pre-training. For impala, vulturine guineafowl (\textit{Acryllium vulturinum}), plains zebra (\textit{Equus quagga}), dikdik, baboon, bird (other), and hare (\textit{Lepus victoriae}), we observed higher accuracy when training under the highest level of ground truth uncertainty, but only in models pre-trained on ImageNet. By contrast, in elephants there was a clear negative effect of lowering the agreement threshold, but only in non-pre-trained models. 

Pre-training on ImageNet resulted in substantial accuracy improvements for most species; in some cases, the pre-trained model performance -- particularly when subsequently trained under high ground truth uncertainty -- matched or exceeded the performance of human volunteers (Fig. \ref{fig:species_heatmap_prickly_pear_balanced}). The greatest improvements resulting from pre-training were observed for dikdik, helmeted guineafowl, and bird (other), for which non-pre-trained accuracy was generally poor. By contrast, for hippopotamus (\textit{Hippopotamus amphibius}) and camels pre-training resulted in little or no improvement. 

For Snapshot Serengeti (Fig. S1B), using the 66\% label agreement threshold resulted in the greatest accuracy for buffalo, hartebeest (\textit{Alcelaphus buselaphus cokii}), elephant, impala, Grant's gazelle, guineafowl (\textit{N. meleagris}), and lions (\textit{Panthera leo}) when the model was pre-trained on ImageNet, and hartebeest, impala, Grant's gazelle, and hippopotamus when trained from scratch. 
By contrast, in the pre-trained models accuracy for hippopotamus and spotted hyena (\textit{Crocuta crocuta}) was highest with a 100\% agreement threshold, and decreased when the threshold was lowered. There was little effect of changing the agreement threshold on wildebeest (\textit{Connochaetes taurinus}), the most abundant class; the second most abundant class, zebra (\textit{E. quagga}), exhibited lower accuracy when the training threshold was lowered in non-pre-trained models, but no clear effect of changing the threshold when the model was pre-trained on ImageNet. 

\subsubsection{Does training on more uncertain images improve generalisation between datasets?}
Comparison between models trained using Snapshot Serengeti with different levels of ground truth uncertainty (66\% or 100\% label agreement threshold), then fine-tuned on Prickly Pear Project Kenya, did not reveal any consistent effect on test performance (Fig. \ref{fig:experiments_entropy}B). While we did observe negative and positive effects of lowering the agreement threshold when the model was subsequently fine-tuned on 66\% agreement and 90\% agreement threshold data respectively, these effects were small and only observed in one entropy bin each, and only for models not pre-trained on ImageNet (Fig. \ref{fig:experiments_entropy}B).  These findings suggest that training under higher ground truth uncertainty does not improve generalisation between datasets. 

We did observe that in models that were not pre-trained on ImageNet, training on Snapshot Serengeti generally resulted in a small accuracy boost relative to training from scratch (Fig. \ref{fig:experiments_entropy}B). This effect was most pronounced for easier test images (entropy = 0.0000 to 0.6469), as well as the hardest images (entropy = 1.2938-1.6172) when the model was fine-tuned under higher ground truth uncertainty (90\% and 66\% agreement thresholds). By contrast, in models pre-trained on ImageNet there was no improvement in accuracy gained by additionally training on Snapshot Serengeti. 

However, when we examined the number of epochs required to obtain a training accuracy of 0.97 for each model, we observed that pre-training on ImageNet followed by training on Snapshot Serengeti greatly reduced the required epochs, compared to training on either ImageNet or Snapshot Serengeti alone (Fig. \ref{fig:experiments_epochs}). Fine-tuning on 100\% agreement threshold data required the fewest epochs; in this case, the number of epochs was further slightly reduced by training on 66\% Snapshot Serengeti data. Training models from scratch was by far the most computationally intensive approach, requiring more than twice as many epochs as any other model (Fig. \ref{fig:experiments_epochs}). In general, training using a lower agreement threshold required more epochs, suggesting a trade-off between computational cost and model performance.

\subsubsection{Does ground truth uncertainty interact with class imbalance?}
\begin{figure}
    \centering
    \includegraphics[width=1\linewidth]{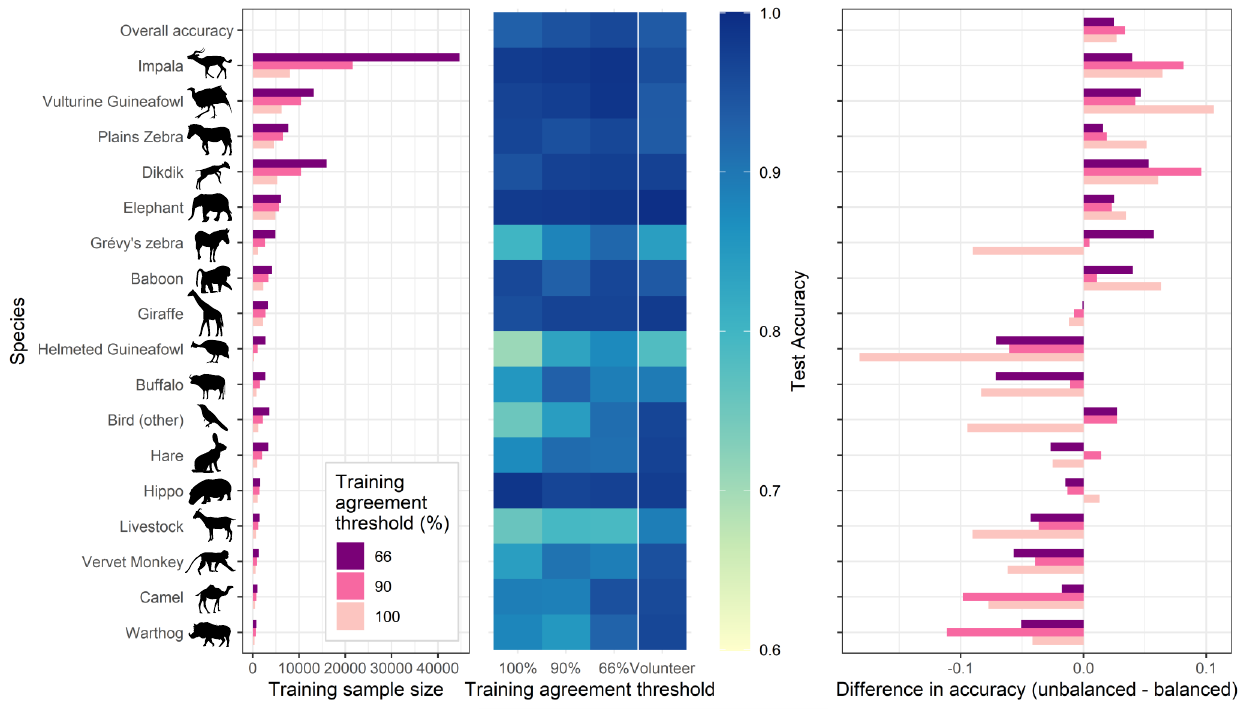}
    \caption{Species-specific classification accuracies for the Prickly Pear Project Kenya dataset, using unbalanced training data. \textbf{Left:} Number of images per species in training set set under different training label agreement thresholds ($\geq$66\%, 90\%, 100\%). \textbf{Centre:} Test accuracy (true positive rate) for different agreement thresholds and human volunteers. The top row displays overall test accuracy across all species. \textbf{Right:} Difference in test accuracy between models trained on unbalanced and balanced data; positive values indicate better performance when trained on unbalanced data. Separate bars for each species represent different label agreement thresholds. Species are ordered by frequency of occurrence in the test set, from most (top) to least common (bottom).}
    \label{fig:unbalanced_results}
\end{figure}

Removing the 700-sample threshold for the Prickly Pear Project Kenya training dataset resulted in the inclusion of an additional 40,752, 74,657, and 118,036 images for the 100\%, 90\% and 66\% agreement threshold datasets respectively. However, removing the threshold also generated a significant degree of class imbalance in all three datasets (Fig. \ref{fig:unbalanced_results}, left panel). 

The degree of imbalance was greatest for the 66\% threshold dataset ($H$ = 2.18), which was dominated by 44,622 impala examples; the next most abundant class was dikdik, which had only 35\% the number of images as impala (Fig. S2). The rarest class, warthog, had only 2\% of the impala images. The 90\% threshold dataset had an intermediate degree of imbalance ($H$ = 2.31), while the 100\% threshold dataset was the most balanced ($H$ = 2.42). In the latter dataset impala was still the most common class, but four other classes (vulturine guineafowl, plains zebra, dikdik, and elephant) had $>$50\% the number of images as impala (Fig. S2). Furthermore, the rarest class in the 100\% dataset -- helmeted guineafowl -- had 4\% of the number of impala examples. 

Similarly to models trained on balanced data (with pre-training on ImageNet), models trained using a lower agreement threshold performed better across all entropy classes, but especially on more difficult test images (Fig. \ref{fig:experiments_entropy}C). Models trained using an unbalanced dataset performed slightly better than balanced models at low to intermediate entropy bins, but often exhibited worse performance on higher entropy images (Fig. \ref{fig:experiments_entropy}C). 

The species-level classification accuracy of models trained on unbalanced data varied among species (Fig. \ref{fig:unbalanced_results}, centre panel). For several species -- including Grévy's zebra (\textit{E. grevyi}), helmeted guineafowl, bird (other), camel, and warthog -- there was a clear increase in performance as the training agreement threshold was lowered. Training on unbalanced data generally improved test accuracy for more common classes, and reduced accuracy for rare classes; the difference in species-level accuracy was generally around $\pm$5-10\%, but varied among species and training agreement thresholds (Fig. \ref{fig:unbalanced_results}, right panel). Relative to the balanced dataset results, classification accuracy was consistently higher for impala, vulturine guineafowl, plains zebra, dikdik, elephant, and baboon. By contrast, classification accuracy consistently decreased for the helmeted guineafowl, buffalo, livestock, vervet monkey, camel, and warthog classes. The other five classes -- Grévy's zebra, bird (other), giraffe (\textit{Giraffa reticulata}), hare, and hippopotamus -- exhibited varying results, with no consistent pattern across the species. Although the difference in accuracy varied among training agreement thresholds, this variation did not appear to follow a clear overall pattern. 

Interestingly, switching to an unbalanced training dataset resulted in the pattern of incorrect classifications for two pairs of easily confused species -- plains and Grévy's zebra, and vulturine and helmeted guineafowl -- changing to more closely resemble the errors made by human volunteers (Figs. S3-4). Whereas models trained on balanced data tended to mistake each species in the pair for the other at similar rates, both models trained on unbalanced data and human volunteers tended to mistake the rarer species for the more common (\textit{i.e.,} classify Grévy's zebra as plains zebra, and helmeted guineafowl as vulturine guineafowl), but not vice versa. 

\section{Discussion}
Where ecological image data are classified by multiple annotators -- such as volunteer citizen scientists -- disagreement among annotators creates uncertainty in the ground truth data used to train deep learning models. We found that training on data with a higher level of ground truth uncertainty improved overall classification accuracy on test images that were more difficult for human volunteers. The effects on species-level accuracy were generally positive; often, but not always, the clearest gains in accuracy were observed in species where volunteer accuracy was relatively poor. Both overall and species-level classification accuracy were enhanced by pre-training on ImageNet. Additionally, in pre-trained models, training under higher ground-truth uncertainty improved test accuracy even for relatively easy images. However, although training under higher ground truth uncertainty generally increased classification accuracy on unseen images from the same dataset, we did not find that it improved generalisation to a different dataset. When we trained models using an unbalanced dataset -- rather than balancing the training data to the same number of examples -- we still observed a clear overall benefit to training under higher ground truth uncertainty, particularly for difficult images. Using an unbalanced training set generally improved classification accuracy for common species, and decreased accuracy for rare species; although the magnitude of this change varied among different levels of ground truth uncertainty, this variation did not exhibit a clear pattern. Our findings suggest avenues for improving the application of deep learning models in the common situation where multiple labels are available.

\subsection{Training under higher ground truth uncertainty improves performance on difficult images} 
Lowering the agreement threshold required for an image label to be accepted -- increasing ground truth uncertainty -- expands the diversity of examples presented to the model, but also increases the chance that incorrect labels are introduced. We found that the net effect was to improve models' overall classification accuracy on test data, especially for images which were more difficult for human volunteers. We also observed improvements in species-level accuracy for several classes where volunteer accuracy was relatively poor, including helmeted guineafowl and buffalo (for Prickly Pear Project Kenya) and hartebeest, impala, and Grant's gazelle (for Snapshot Serengeti). In camera trap studies, difficult images often fundamentally differ from easier images. For example, difficult images often include partial captures where key diagnostic features may be obscured, animals far from the camera, or images with poor lighting (Fig. \ref{fig:image_examples}). Consequently, exposure to difficult images during training allows deep learning models to gain experience with a greater diversity of examples, improving classification accuracy during testing. Like camera trap images, difficult examples in other ecological image data often contain features that distinguish them from easier examples. For instance, plant images from herbaria can be difficult to classify because they are poor quality or display unusual morphological variation \parencite{shirai_development_2022}, while difficult acoustic spectrograms may contain overlapping sounds or background noise \parencite{napier_advancements_2024}. Consequently, we expect that allowing deep learning models to encounter more difficult images, by training under higher ground truth uncertainty, will improve classification accuracy across a wide range of image types in ecology.

In our study, ground truth uncertainty was generated by multiple annotators. Although this situation is applicable to a wide range of applications in ecology, in other cases only a single label is available. In these cases, it may be possible to use model-free image complexity metrics (\textit{e.g.,} \cite{cgscore, mahon2024}) to quantify example difficulty \parencite{Kwok2024} without the need for multiple annotators. If training on examples with high complexity metric scores produces similar benefits to training under higher ground truth uncertainty, then these benefits would be available across all ecological image datasets rather than just those with multiple annotations. Consequently, we suggest that this represents a key avenue for future research. 

\subsection{The benefits of ground truth uncertainty are enhanced by pre-training}
When models were pre-trained on ImageNet, the benefits of training under higher ground truth uncertainty were evident even for relatively easy images. By contrast, in models trained from scratch, training on higher-uncertainty images only improved performance on intermediate to high-difficulty images. This benefit of pre-training occurred in addition to improved accuracy and a reduction in the number of epochs required for training, which are expected benefits of pre-training \parencite{Yosinski2014, He2019}. Notably, models which were pre-trained on ImageNet outperformed those pre-trained on Snapshot Serengeti, even though the latter is closer in style and content to the test data. Furthermore, pre-training on both ImageNet and Snapshot Serengeti provided no accuracy benefit over pre-training on ImageNet alone, although models pre-trained on both datasets did take fewer epochs to fine-tune. In computer vision, pre-training is thought to enable models to learn generic features of images (\textit{e.g.,} edges, textures), supporting performance on subsequent image-related tasks \parencite{He2019}. The size and diversity of the ImageNet dataset (over a million images representing 1000 classes) appears more useful for this purpose than the relatively small and homogeneous Snapshot Serengeti. Consequently, we suggest that practitioners use models which are pre-trained on generic image datasets, such as ImageNet, and fine-tune them on their own data to maximise classification accuracy and reduce training time. Although we did not find that additional pre-training on Snapshot Serengeti improved accuracy, it reduced the computational cost of fine-tuning. This suggests that wider re-use of models could reduce the environmental footprint associated with the application of deep learning in ecology. 

\subsection{Training under ground truth uncertainty does not aid transfer to novel contexts}
The failure of models to generalise to novel contexts (e.g., sites, study systems) is a persistent concern in the application of deep learning models to ecological image data \parencite{Beery19, shepley_automated_2021}. We found that models pre-trained on Snapshot Serengeti using different label agreement thresholds, and subsequently fine-tuned and tested on Prickly Pear Project Kenya, did not exhibit differences in performance. This suggests that training on more uncertain images does not improve generalisation between datasets. Poor generalisation is thought to be caused by models overfitting to irrelevant features of images, such as backgrounds \parencite{Beery2018, shepley_automated_2021}. Although higher-uncertainty images may differ in these features to some degree (\textit{e.g.,} where the background is obscured by poor visibility), any increase to background diversity is apparently not sufficient to mitigate this overfitting. An increasingly common approach for improving generalisation in camera trap data is to use an analysis pipeline in which an object detection model is first used to produce bounding boxes for animals in an image (regardless of species), which are subsequently cropped and passed to a classifier \parencite{gadot_crop_2024}. We expect that our main finding -- that training on higher-uncertainty improves performance, especially on difficult images -- will transfer to this pipeline, because the model will still benefit from encountering a greater diversity of examples. One potential difficulty is that object detectors may perform poorly on high-difficulty images, \textit{e.g.,} failing to place the bounding box around the animal. We suggest that exploring the role of ground truth uncertainty in pipelines which incorporate object detection is an important topic for future research. 

\subsection{Class imbalance affects species-level performance, but benefits of ground truth uncertainty are still evident}
When we trained models using an unbalanced dataset -- greatly increasing the sample size for several species, while simultaneously introducing a substantial degree of class imbalance -- we still observed that training under higher ground truth uncertainty improved performance on difficult images. We also found that class imbalance affected species-level accuracy, generally improving accuracy for common species but harming accuracy for rare species, in agreement with the expected effect of class imbalance (\cite{prince_understanding_2024}, p.73). In addition, the level of class imbalance was greater when higher-uncertainty images were included in the training set. This suggests that whether increasing the number of high-uncertainty examples by including all available images, rather than using a balanced subset, is the best strategy depends on whether accuracy for common or rare classes is more important. 

We also observed an additional effect of training using an unbalanced dataset -- the pattern of misclassification between easily-confused pairs of species changed to more closely resemble errors made by human volunteers. Specifically, both unbalanced models and volunteers tended to make directional errors in which rare species were mistaken for common species (Grévy's zebra for plains zebra, and helmeted guineafowl for vulturine guineafowl), whereas models trained on balanced data did not. This finding has important implications for approaches which combine human and deep-learning classifications, such as combined consensus (\textit{e.g.,} \cite{sharpe25}) or human-in-the-loop approaches (\textit{e.g.,} \cite{norouzzadeh2021}); these combined deep learning and citizen science approaches have great potential to increase accuracy for camera-trap monitoring while promoting public engagement \parencite{green20,adam21,sharpe25}. 
Combined approaches may perform better if humans and models make different mistakes, as areas of disagreement can then be identified for further validation. By contrast, if the mistakes made by humans and models are similar, then errors may enter the combined classifications undetected. 

\subsection{Conclusions and recommendations}
We found that training deep learning models under higher ground truth uncertainty improved classification accuracy on test data, particularly for images which were difficult for human volunteers, but did not improve generalisation to novel contexts. Class imbalance improved classification accuracy for common species and decreased accuracy for rare species. However, class imbalance did not exhibit a clear interaction with ground truth uncertainty, and the benefits of training under higher uncertainty remained evident. 

Our findings have implications for the application of deep learning models not only to camera trap data, but  across a wide range of ecological image types where multiple labels are available. We propose that practitioners can maximise accuracy by including labels with a moderate level of disagreement in the training data, and by using models which have been pre-trained on general image datasets (\textit{e.g.,} ImageNet). Using pre-trained models also substantially reduces the computational cost (number of epochs) required to train models. We suggest that the choice between class-balanced and unbalanced training data depends on whether accuracy for rare or common classes is more important.

Deep learning and volunteer citizen scientists have both made substantial contributions to the analysis of ecological image data. As deep learning is integrated into more ecological workflows, and classification accuracy continues to improve as architectures are developed and refined, it is important to consider the role of human volunteers in future research projects. We believe that there is great potential in combined workflows that include both volunteer and deep learning classifications. Our study has illustrated that this integration can be improved by leveraging the classifications of multiple volunteers to provide information about example difficulty. Furthermore, we have illustrated fundamental differences in the types of mistake made by humans and models, which could be exploited to produce more accurate combined classifications. Including human classifiers also creates unique benefits. For example, humans can generate insights -- such as identifying interesting behaviours, species interactions, or other occurrences -- that lie beyond the scope of the model's task. Finally, engaging people in ecological research provides a range of benefits including improved wellbeing, skills development, and connection to nature \parencite{green20}, which are important to preserve.

\printbibliography[title={References}]

\end{document}


\maketitle

\renewcommand{\thefigure}{S\arabic{figure}} \setcounter{figure}{0}

\begin{figure}
    \centering
    \includegraphics[width=1\linewidth]{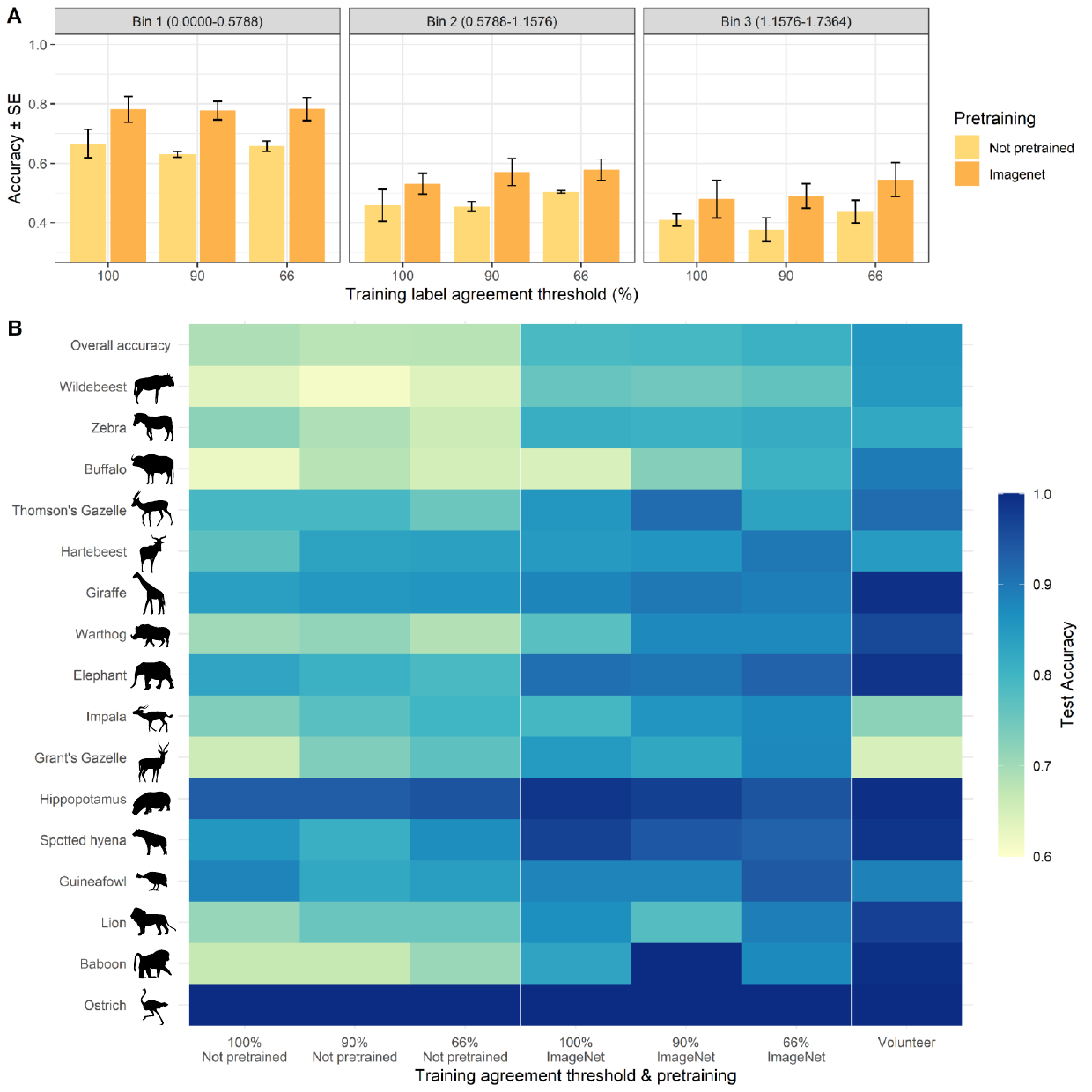}
    \caption{Results for the Snapshot Serengeti test set, for models trained under different volunteer agreement thresholds (\%) for label acceptance and either not pre-trained or pre-trained on ImageNet. \textbf{A)} Test accuracy (proportion of correct classifications; mean $\pm$ standard error) across different levels of image difficulty as measured by the Shannon entropy of the volunteer classifications; higher entropy values (towards right) indicate more volunteer disagreement, and hence more difficult images. Sample sizes for entropy bins are (left to right) 2635, 298, 49. Three images with entropy $>1.7364$ were excluded. \textbf{B)} Species-level classification accuracy (true positive rate). Volunteer accuracy is also displayed on the rightmost column. The top row displays overall test accuracy across all species. Species test set sample sizes are: wildebeest (1356), zebra (782), buffalo (164), Thomson's gazelle (151), hartebeest (97), giraffe (79), warthog (75), elephant (73), impala (72), Grant's gazelle (46), hippopotamus (25), spotted hyena (23), guineafowl (17), lion (15), baboon (8), ostrich (2). 
  }
    \label{fig:supp_serengeti_results}
\end{figure}

\begin{figure}
    \centering
    \includegraphics[width=1\linewidth]{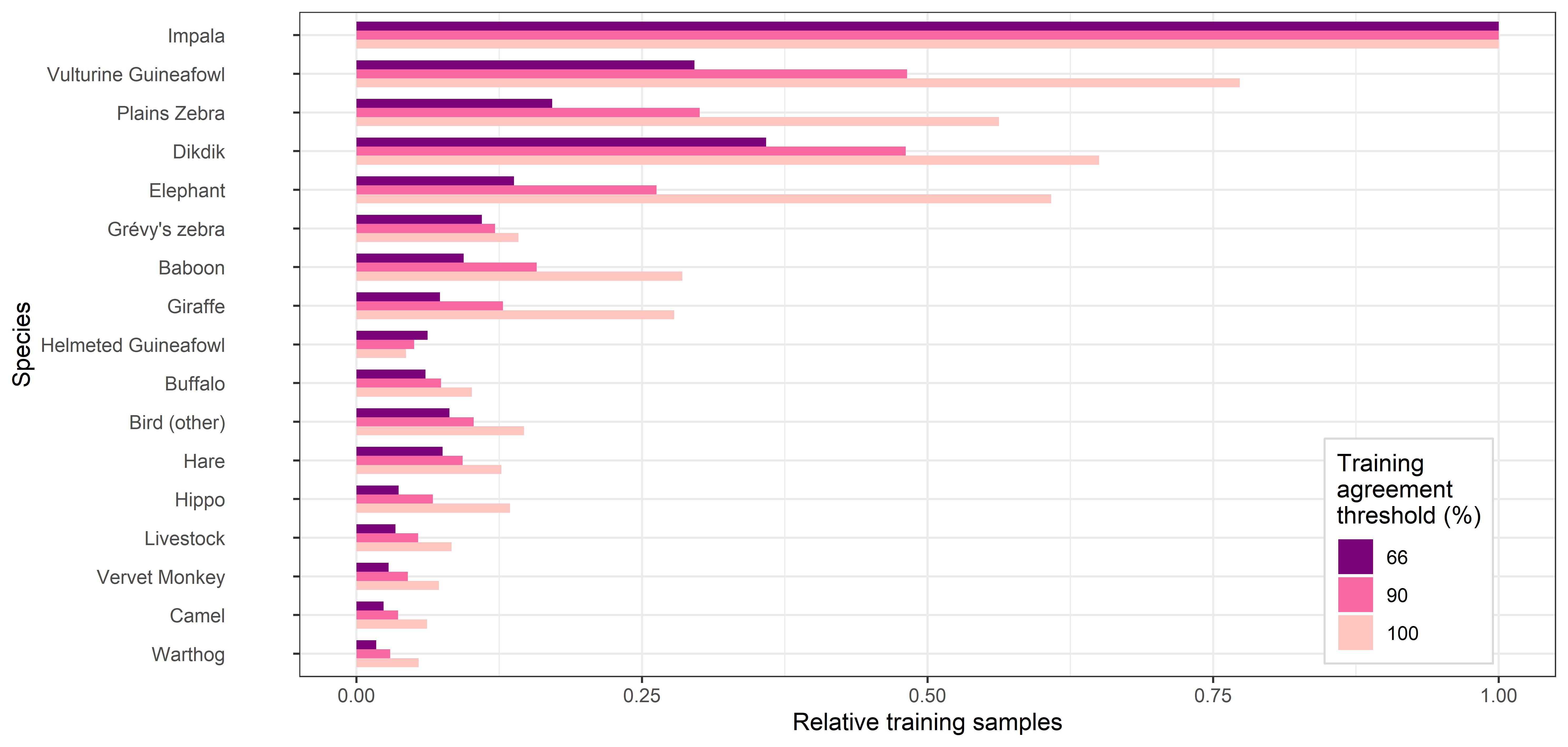}
    \caption{Relative number of training samples for the unbalanced Prickly Project Kenya dataset; number of samples for each class is expressed as a proportion of the most abundant class for each training agreement threshold.}
    \label{fig:supp_unbalanced_class_prop}
\end{figure}

\begin{figure}
    \centering
    \includegraphics[width=1\linewidth]{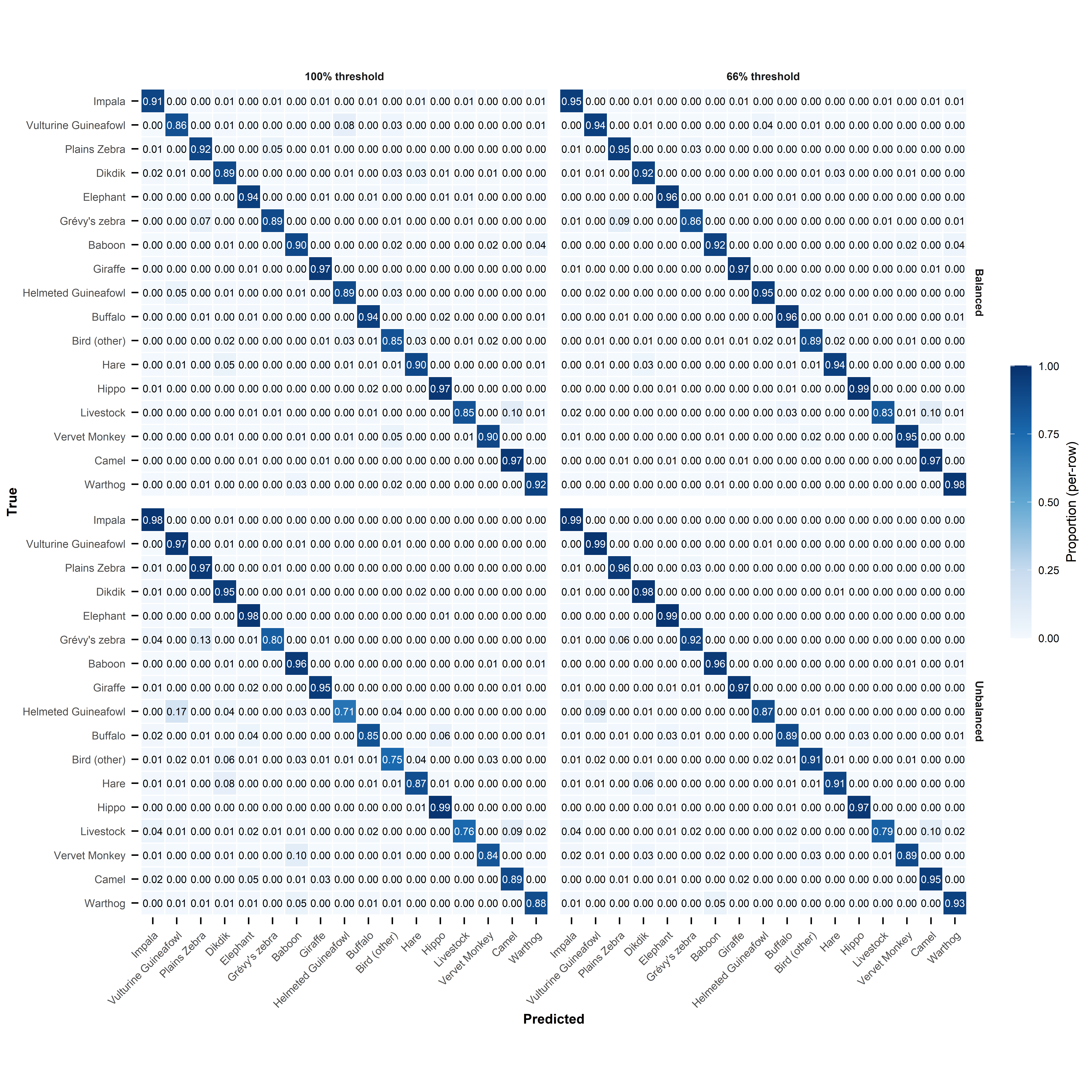}
    \caption{Confusion matrices for models trained using 100\% (left column) or 66\% (right column) label agreement thresholds, and balanced (top row) or unbalanced (bottom row) Prickly Pear data training sets. Matrix rows represent true test set labels, while matrix columns represent model predictions. Species are sorted by the number of (true) occurrences in the test set.}
    \label{fig:supp_cmat_models}
\end{figure}

\begin{figure}
    \centering
    \includegraphics[width=1\linewidth]{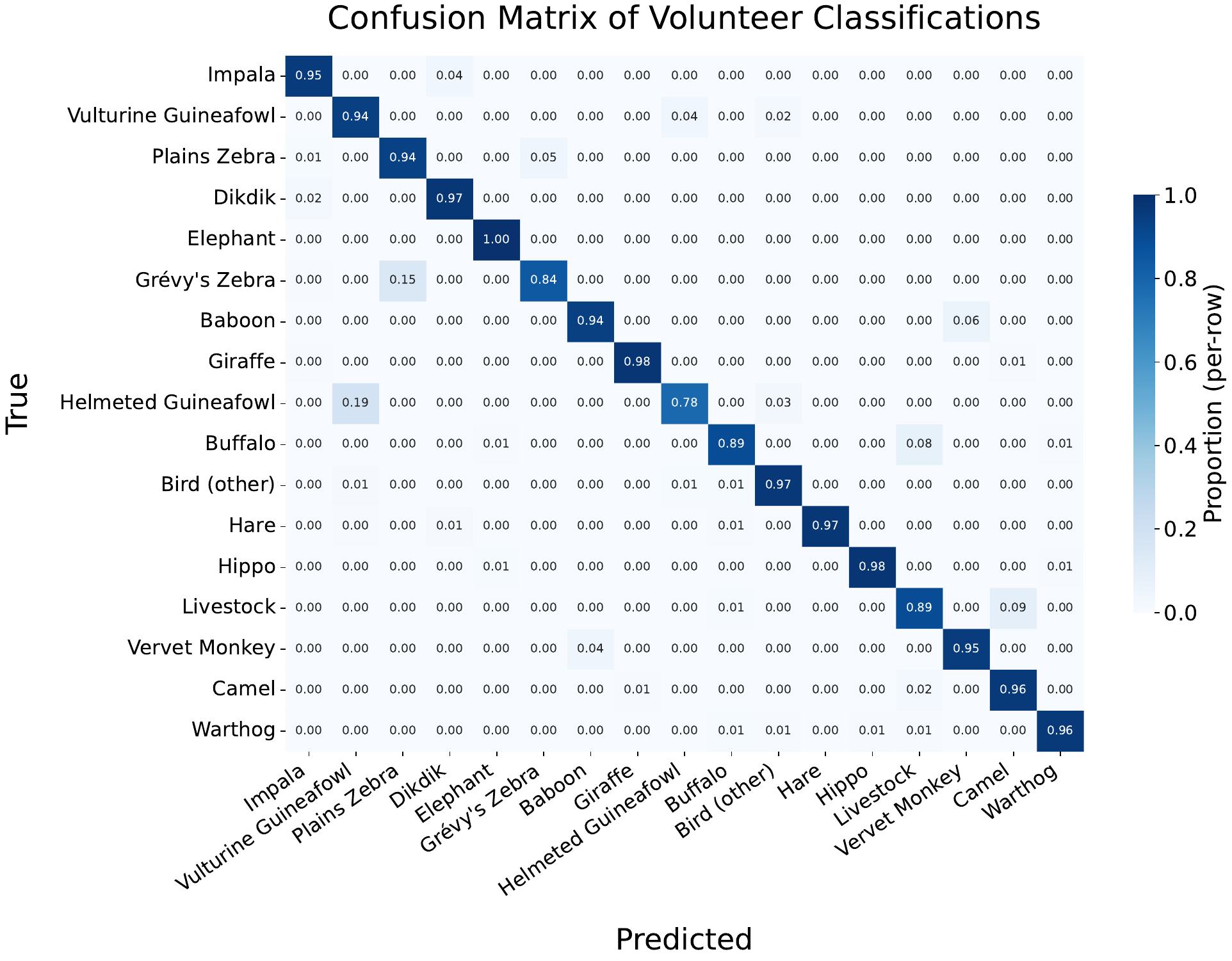}
    \caption{Confusion matrix for human volunteers for the Prickly Pear dataset. Matrix rows represent true test set labels, while matrix columns represent volunteer predictions. Species are sorted by the number of (true) occurrences in the test set.}
    \label{fig:supp_cmat_volunteer}
\end{figure}